\pdfoutput=1
\documentclass{bmvc2k}

\usepackage{amssymb}
\usepackage{mathtools}
\usepackage{lipsum}
\usepackage{wrapfig}

\usepackage[ruled,linesnumbered]{algorithm2e}

\title{Improving Weakly-Supervised Object Localization By Micro-Annotation} 

\addauthor{Alexander Kolesnikov}{akolesnikov@ist.ac.at}{1}
\addauthor{Christoph H. Lampert}{chl@ist.ac.at}{1}

\addinstitution{
 IST Austria\\
 Klosterneuburg, Austria
}

\runninghead{A. Kolesnikov and C. H. Lampert}{Improving Object Localization}

\def\eg{\emph{e.g}\bmvaOneDot}
\def\ie{\emph{i.e}\bmvaOneDot}

\begin{document}

\maketitle

\begin{abstract}
Weakly-supervised object localization methods tend to fail for object 
classes that consistently co-occur with the same background elements, 
\eg \emph{trains} on \emph{tracks}. 
We propose a method to overcome these failures by adding a very 
small amount of model-specific additional annotation. The main 
idea is to cluster a deep network's mid-level representations 
and assign \emph{object} or \emph{distractor} labels to each cluster. 
Experiments show substantially improved localization results on 
the challenging ILSVC2014 dataset for bounding box detection and 
the PASCAL VOC2012 dataset for semantic segmentation.
%
\end{abstract} 

\begin{wrapfigure}{r}{0.49\textwidth}
    \vskip-\baselineskip
    \includegraphics[width=0.48\textwidth]{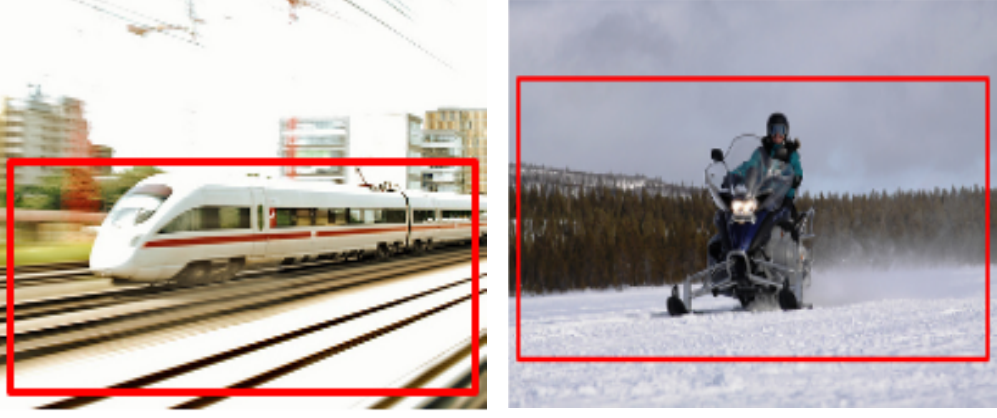}
  \caption{Failure cases of weakly-supervised object localization: based only on per-image class labels one 
  cannot distinguish between objects (\emph{train, snowmobile}) and consistently co-occurring distractors (\emph{tracks, snow}).}\label{fig:page1teaser} 
\end{wrapfigure}

\section{Introduction}

\begin{figure}[t]
        \center
        \includegraphics[width=0.85\textwidth,height=0.33\textheight]{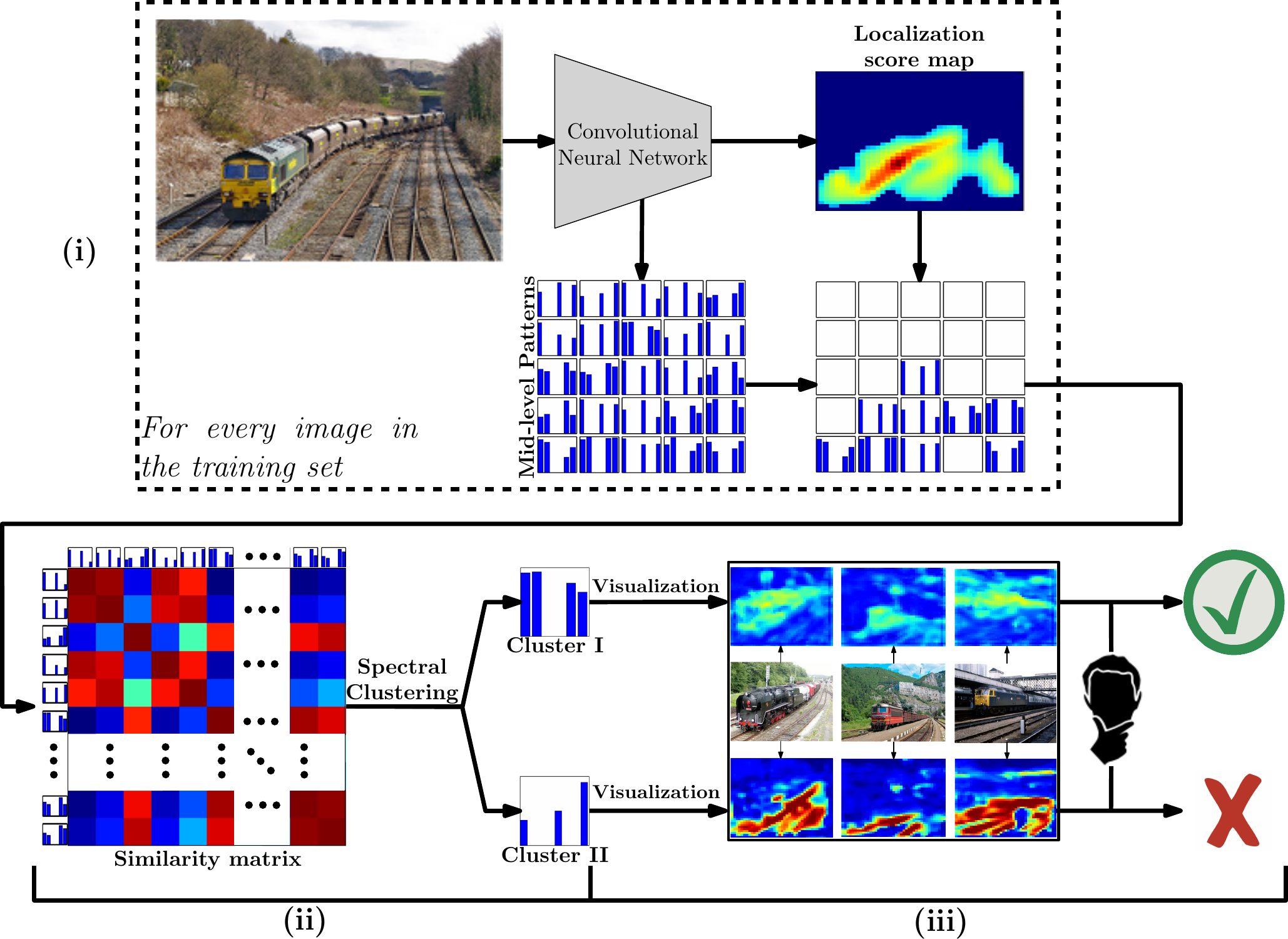}
        \caption{Schematic illustration of the proposed micro-annotation approach:
                 (i) for every image region across all training images that is predicted to show the object 
                 of interest (here: \emph{train}), compute its mid-level feature representation (pattern) and form 
                 a pool of patterns,
                 (ii) find characteristic clusters in this pool by spectral clustering,
                 (iii) visualize the clusters by heatmaps and ask a user to annotate them as representing either 
                 the object or a distractor.
                 %
         } 
        \label{fig:method}
\end{figure}

A crucial step for automatic systems for image understanding 
is the ability to localize objects in a scene in the form 
of bounding boxes or segmentation masks.
Over the last decade, computer vision research has made great 
progress on this task by employing machine learning techniques, 
in particular deep learning. 
However, learning-based methods require large amounts of annotated 
training data, and manually annotating images with object locations 
is a tedious and expensive process.
It is therefore an important, and largely unsolved, problem to develop 
object localization methods that can be trained from weak supervision, 
by which in the context of this work we mean per-image class labels 
or tags.
Analyzing the existing method in this field, it becomes apparent that 
certain object classes, for example \emph{trains} or \emph{boats}, 
are consistently harder to localize from weak supervision than others, 
in the sense that there is a big gap between the prediction quality 
of the models trained from just per-image class labels and the 
quality achieved by models trained with full supervision.

One major reason for these failure cases is illustrated in 
Figure~\ref{fig:page1teaser}.
Weakly-supervised object localization tends to fail when object 
classes systematically co-occur with \emph{distractors} (background or 
certain other classes), for example \emph{trains} with \emph{tracks}. 
Per-image class annotation simply does not contain necessary 
information 
to reliably learn the difference between objects and distrators.
In this work, we argue that the best way to overcome this problem 
is to collect a tiny amount of additional annotation, which we call \emph{micro-annotation}.

%
%
The approach relies on hypothesis that even though it might 
be impossible for a classifier to learn from weak per-image 
annotation which parts of an image are the object of interest 
and which are distractors, it will still be possible to 
distinguish both groups from each other by clustering their 
appearance in a suitable representation. 
Then, all we need in order to improve the quality of a 
weakly-supervised object localization systems is a way to 
find out which clusters belong to distractors -- which is an 
easy task for a human annotator -- and suppress them.

This micro-annotation approach can be used in combination 
with many existing localization methods. 
In this work we combine it with the current state-of-the-art 
methods for weakly-supervised bounding box prediction and 
for weakly-supervised semantic segmentation, showing improved 
results on the challenging ILSVR2014 and PASCAL VOC2012 datasets.

Apart from its practical usefulness, an interesting aspect of 
this approach is that it asks for user annotation after an 
initial model has already been trained. 
This allows the requested information to depend on the original 
model's output, hopefully with the effect that the new 
information has a maximally beneficial effect on the prediction 
quality.
%
%
This setup resembles \emph{active learning}, but with an even better 
relation between the amount of annotation and the model improvement.
In active learning, the annotation provides information through 
the labeling of individual images, and each provided label 
typically influences the model parameters by an amount inversely 
proportional to the total size of the training set. 
Consequently, active learning is most beneficial for models 
trained on small datasets.
In our approach, a single user interaction can have a large effect 
on the model parameters and thereby the prediction quality, namely 
when it establishes that all detected patterns of a certain type 
are distractors and should be suppressed. 
The size of the training set plays no role for this effect, and 
indeed we observe a significant improvement even for models 
trained on very large datasets. 
%


%
%
%
%

\vspace{-4mm}
\section{Related work}
Many methods for object localization have been proposed that 
can be trained in a weakly-supervised way from per-image class label annotation.
The majority of these methods predict either object bounding boxes~\cite{bilen2014weakly,bilen2015weakly,cinbis2014multi,deselaers2010localizing,song2014learning,song2014weakly,wang2015large,
                                               zhou2015cnnlocalization,bazzani2016self}
or per-pixel segmentation masks~\cite{vasconcelos2006weakly,verbeek2007region,
                                                 vezhnevets2010towards,vezhnevets2011weakly,vezhnevets2012weakly,
                                                 xu_cvpr2014,xu2015learning,zhang2015weakly,
                                                 pathak2014fully,pathak2015constrained,pinheiro2015image,papandreou2015weakly,
                                                 kolesnikov2016seed}.
The currenlty most successful approaches obtain localization hints from 
convolutional neural networks that are trained for the of image classification, \eg~\cite{zhou2015cnnlocalization,kolesnikov2016seed}.
The micro-annotation method can in principle be used on top of any of 
such method, as long as that has the ability to produce per-location 
score maps. 

Much fewer works have studied how the process of data annotation 
can be improved by the power of strong computer vision systems.
Two related research directions are \emph{learning with humans in the loop}~\cite{branson2010visual} 
and \emph{active learning}~\cite{activelearning}.
In the human in the loop concept, an automatic system and a human 
user work together during the prediction stage. For example, in 
a fine-grained classification task~\cite{branson2010visual,deng2013fine,wah2011multiclass,wah2014similarity}, the machine would output a 
selection of possible labels and the human user would pick the 
most appropriate one. 
Typically, this process does not improve the model parameters, 
though, so feedback from a human is always required to make
high quality predictions.
Active learning also has the goal of harvesting human expertise, 
but it does so during the training stage: an automatic system 
has access to a large number of unlabeled images and can ask a 
human annotator to specifically annotate a subset of 
them~\cite{Freytag14selecting,joshi2009multi,kapoor2007active,qi2008two}. %
This procedure can reduce the amount of necessary annotation, \eg 
when the system only ask for annotation of images that it is 
not already certain about anyway. 
In the context of object localization, a more efficient weakly-supervised
variant of active learning has been proposed in which a human user 
only has to annotate if a predicted bounding box is correct or not, 
instead of having to draw it manually~\cite{papadopoulos2016we}.
Our approach differs from these setting in particular in the fact that 
we do not ask a user to provide feedback about individual images, but 
about clusters in the learned data representation, which reflects 
information extracted from the whole training set. Thereby, we require 
much less interaction with the annotator, namely in the order of 
the number of classes instead of in the order of the number of 
training images. 

On the technical level, our method is related to recent approaches 
for discovering object detectors in deep convolutional neural 
networks~\cite{quoc2012icml,zhou2014object,simon2015neural}. 
However, these rely on the assumption that part-detectors correspond 
to individual convolution filter outputs, whereas our clustering 
approach finds co-occurring patterns in the distributed representation 
learned by the network. 
The fact that our method identifies image structures by clustering 
visual representations across many images resembles image 
co-segmentation~\cite{rother2006cosegmentation}. 
It differs, however, from these earlier works in how it organizes 
the clustering process and how it uses the structure that are found.

\section{Improving Object Localization By Micro-Annotation}

\begin{figure}[t]
        \center
        \includegraphics[width=0.9\textwidth]{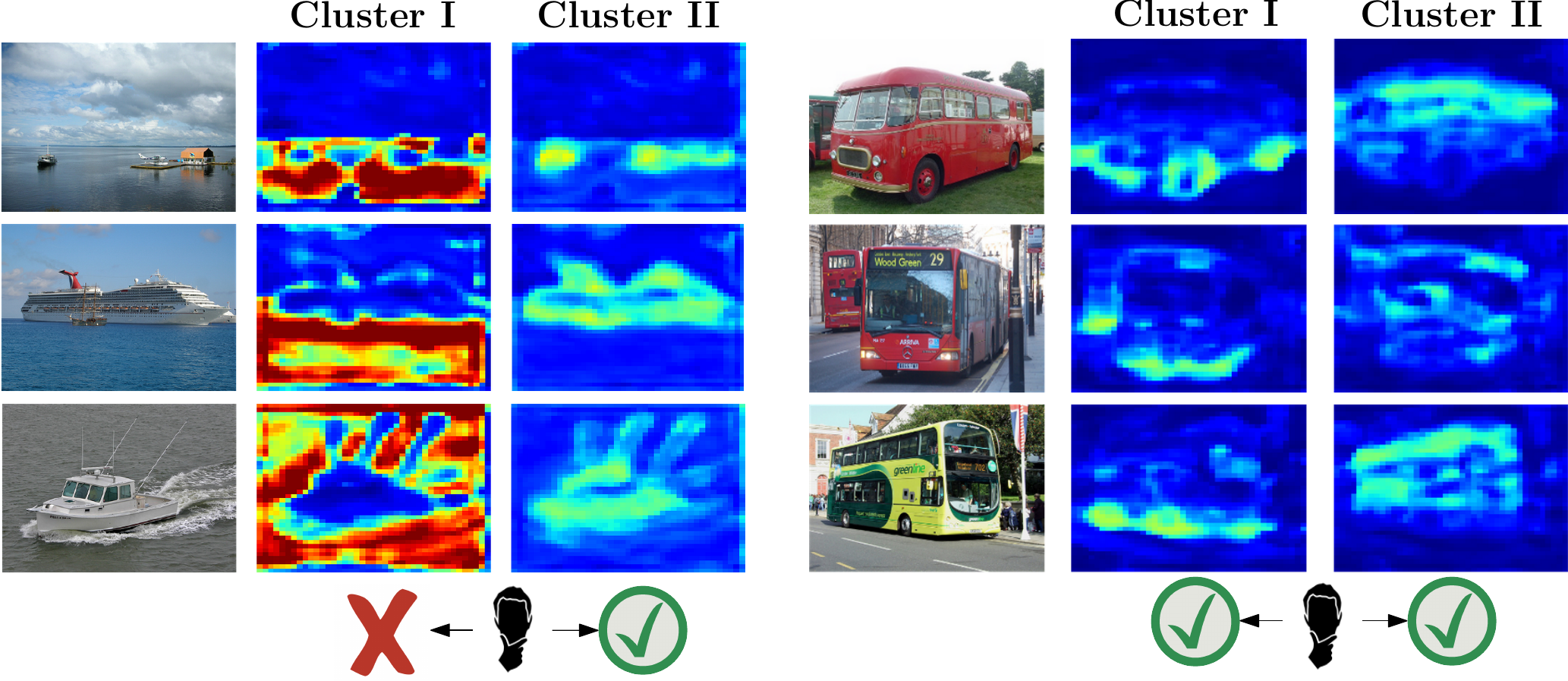}
        \caption{The schematic illustration of the annotation process.
                 For any semantic category, we visualize corresponding mid-level feature clusters by heatmaps.
                 An annotator marks every class as either representing an object of interest or background.}\label{fig:annotation}
\end{figure}

\begin{figure}[t]
        \center
        \includegraphics[width=0.9\textwidth]{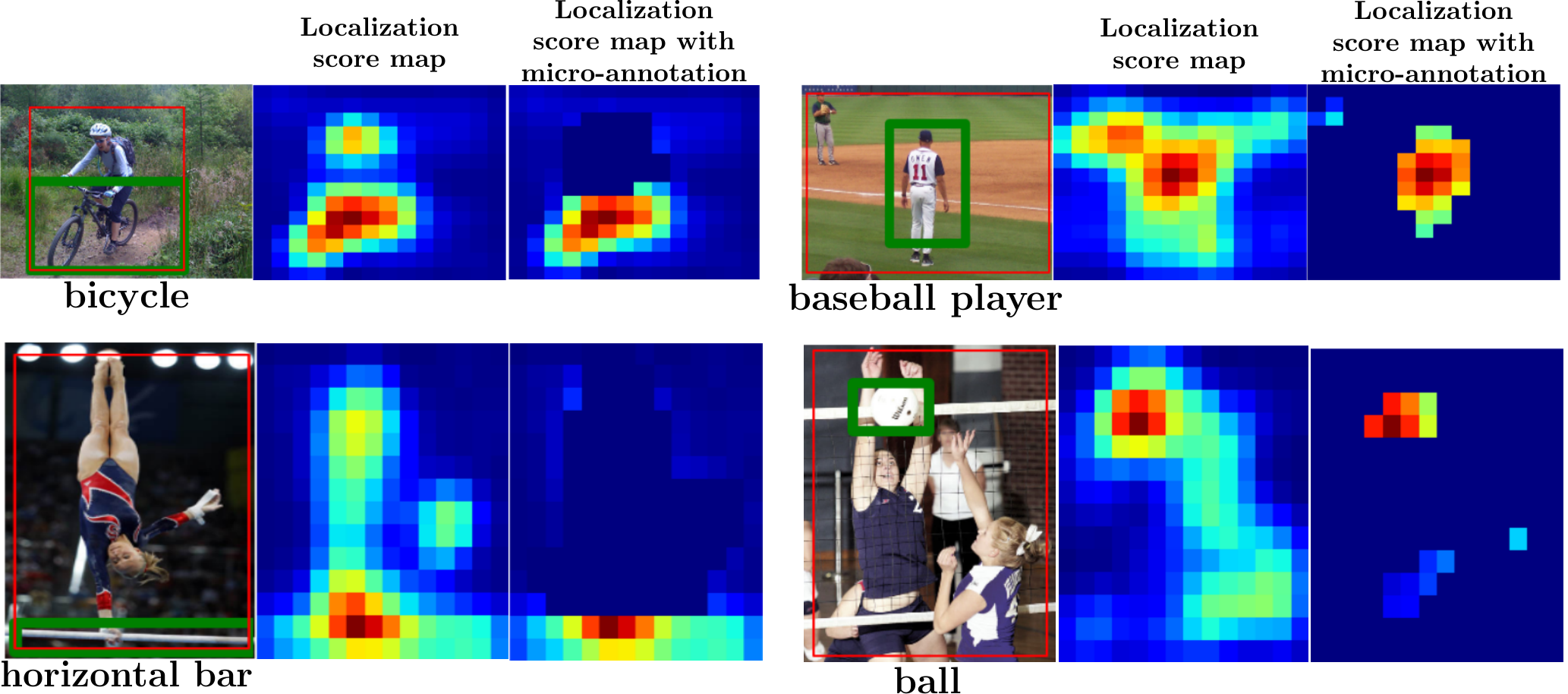}
        \caption{The effect of discarding localizations that correspond to a mid-level representation (pattern) that 
                 is assigned to a cluster annotated as distractor.}\label{fig:test-time}
\end{figure}

In this section we formally introduce the proposed procedure for 
collecting micro-annotation (illustrated in Figure~\ref{fig:method}) 
and improving object localization (illustrated in 
Figure~\ref{fig:test-time}). 
The main steps for obtaining the additional annotation
for each class are: (i) represent all predicted foreground regions of all 
images by feature vectors, (ii) cluster the feature vectors 
(iii) visualize the clusters and let an annotator select which 
ones actually corresponds to the object class of interest.
The information about clusters and their annotation can then
be used to better localize objects: (iv) for any (new) image, 
predict a foreground map using only the image regions that 
match clusters labeled as 'object'.

In the rest of the section we explain these steps in detail. 
For this, we denote the set of training images by $\mathcal{D}$ and 
assume a fixed set of semantic categories $\mathcal{Y}$ that we 
want to localize. 
The subsequent construction can be performed independently for each 
object class. By $y\in\mathcal{Y}$ we always denote the current 
class of interest. 

We assume that we are given a pretrained deep convolutional neural 
network, $f$, that predicts the presence of semantic categories 
for an input image $X$, but that can also be leveraged to predict 
the spatial location of semantic objects in input images.
Formally, we assume that each image is split into a set of (rectangular) 
regions, $\mathcal{U}$, and that $f$ gives rise to a \emph{scoring function}
that assigns a localization score, $S^y_u(X)$, to each image region 
$u \in \mathcal{U}$.
Furthermore, we assume the availability of a \emph{thresholding procedure} 
that converts \emph{score maps}, $S^y(X) \in \mathbb{R}^{|\mathcal{U}|}$, 
to a set of image regions, $D^y(X) \subset \mathcal{U}$, that represent 
the predicted localization of the class $y$.
%
%
We discuss particular choices for the functions $S$ and $D$, in Section \ref{sec:exp}.

\medskip\noindent\textbf{(i) Region representations.}
At any fixed layer of the deep network we can form a feature vector, 
$\phi_u(X) \in \mathbb{R}^k$, (called a \emph{pattern}) for any region, 
$u \in \mathcal{U}$, by concatenating the real-valued convolutional filter 
activations from the fixed layer. 
For simpler use in the clustering step we assume that $\phi_u(X)$ 
has nonnegative entries, \eg after a ReLU operation, and that it 
is $L^2$-normalized.
These are not principled restrictions, though. Arbitrary features 
could be used in combination with a different clustering algorithm.
We form a \emph{pattern set} $A^y = \{\phi_u(X) | \forall u \in D^y(X), \forall X \in \mathcal{D}\}$, 
\ie the features of all image regions with predicted label $y$.


\medskip\noindent\textbf{(ii) Clustering.} 
We partition the set $A^y$ into a group of clusters, $P^y=\{C_1,\dots,C_k\}$,
by \emph{spectral clustering}~\cite{luxburg2007tutorial}. 
The number of clusters is determined automatically using the fact that 
the $k$-th eigenvalues of the graph Laplacian (computed during the 
clustering) reflects the quality of creating $k$ clusters (Algorithm~\ref{alg:spectral}, line 5).  

General spectral clustering does not scale well to large datasets, because when used 
with a generic similarity measure it requires memory quadratic and runtime cubic in 
the number of patterns. 
This is not the case for us: we use a linear (inner product) similarity measure 
between patterns, which allows us to avoid storing the quadratically sized 
similarity matrix \emph{(line 1)} explicitly. The necessary eigenvalue problem \emph{(line 4)}
we solve efficiently by the Lanczos method~\cite{lanczos1950iteration}, which 
requires only low-rank matrix-vector multiplications. The resulting algorithm 
scales linearly in the number of patterns to be clustered and can thereby be 
applied even to datasets with millions of patterns. 

\begin{algorithm}[t]
    \SetKwInOut{Input}{input}
    \SetKwInOut{Output}{output}

    \Input{patterns $A^y$, eigenvalue threshold $\rho$ (default: $0.7$), lower bound $m$ (default: 2) 
    and upper bound $M$ (default $4$) for the number of clusters}

    Compute the similarity matrix: $W^y \in \mathbb{R}^{|A^y| \times |A^y|}$ with $W_{a,b}^y = \langle a, b \rangle \ge 0$ for all $a, b \in A^y$.

    Compute the diagonal matrix: $D^y \in \mathbb{R}^{|A^y| \times |A^y|}$ with $D_{a,a}^y = \sum_{b \in A^y} W_{a,b}^y$ for all $a \in A^y$.

    Compute the Laplacian matrix: $L^y = D^y - W^y$.

    Compute the $M$ smallest eigenvalues $\{\lambda_1, \dots \lambda_M\}$ and eigenvectors $\{v_1, \dots v_M\}$
    of $(D^y)^{-1}L^y$ by solving the generalized eigenvalue problem $L^y v = \lambda D^y v$.

    Set the number of clusters, $k$, as the number of eigenvalues below $\rho$ or the lower bound.\!\!\!

    Construct matrix $U = [v_1\,| \cdots\,| v_k] \in \mathbb{R}^{|A^y| \times k}$.

    Let $u_a$ be the row of $U$ that corresponds to a pattern $a$.
    
    Use \emph{$k$-means} to cluster the matrix rows $\{u_a\}_{a \in A^y}$ into clusters, $\{U_1, \dots, U_k\}$.

    \Output{clustering $P^y = \{C_1, C_2, \dots, C_k\}$, where $C_i = \{a | u_a \in U_i\}$ for $i=1,\dots,k$}

    \caption{Spectral clustering algorithm}\label{alg:spectral}
\end{algorithm}

\medskip\noindent\textbf{(iii) Cluster visualization and annotation.}
Our main assumption is that any cluster, $C\in P^y$, will
correspond either to (part of) the object of interest, 
or to a distractor. 
To identify which of these possibilities it is, we 
introduce an efficient annotation step.

For each class, we randomly sample a small number, \eg 12, 
of images from the training set. 
For each sampled image $X$ we produce heatmaps, $H^y(X | C)\in\mathbb{R}^{|\mathcal{U}|}$, 
for each cluster, $C$, that depict for each region $u$ 
the average similarity of the region pattern $\phi^y_u(X)$ 
to the patterns in the corresponding cluster, \ie 
\begin{equation}
H^y_u(X | C) = \frac{1}{|C|} \sum_{a \in C} \langle \phi^y_u(X), a\rangle.
\end{equation}
Note that in practice we can compute this value without always summing 
over all patterns: we pre-compute the average cluster pattern, $a_C=\frac{1}{|C|} \sum_{a \in C}a$, 
and use $H^y_u(X | C) =\langle \phi^y_u(X),a_C\rangle$.
We display the heatmaps and ask a human annotator to mark which of 
the clusters correspond to the object class of interest in the images,
see Figure \ref{fig:annotation} for an illustration of the process.
Overall, the annotation requires just one user interaction (a few mouse clicks) 
per class. Our experience shows that each interaction takes in the order 
of a few seconds, so a few hundred classes can be annotated within an hour.

\medskip\noindent\textbf{(iv) Improved object class localization.}
The per-class annotations allow us to obtain better location predictions 
without having to retrain the network. 
Let $X$ be a new image with predicted localization score map $S^y_u(X)$.
For each location $u \in \mathcal{U}$ we determine the cluster that 
results in the highest heatmap score, $C^*  = \min_{C \in P^y} H^y_u(X | C)$. 
If $C^*$ was not annotated as an 'object' cluster, we set $S^y_u(X)$ to $-\infty$,
in order to prevent that the class $y$ will be predicted at the location $u$.
%

\section{Experiments}\label{sec:exp}

In this section we evaluate the effectiveness of our approach. 
We apply it to state-of-the-art methods for predicting object 
bounding boxes and semantic segmentations from image-level 
supervision and report experimental result on two challenging 
computer vision benchmarks: ILSVRC 2014 and PASCAL VOC 2012.

\medskip\noindent\textbf{Predicting object bounding boxes.}
We follow the protocol of the ILSVRC 2014 \emph{classification}-\emph{with}-\emph{localization} challenge:
the goal is to predict which of 1000 object classes is present in an image 
and localize it by predicting a bounding box~\cite{ILSVRC15}.
By the challenge protocol, up to five classes and their bounding boxes can be predicted, 
and the output is judged as correct if a bounding box of the correct class is predicted 
with an intersection-over-union score of at least 50\% with a ground-truth box. 

In this work we are particularly interested in the weakly-supervised 
setting, when models are trained using only per-image category 
information. 
We build on the state-of-the art technique GAP~\cite{zhou2015cnnlocalization}, 
which uses a deep convolutional network with modified VGG~\cite{simonyan2014very} 
architecture.
Internally, GAP produces localization score maps, $S^y(X)$, by means of the 
\emph{CAM (class activation maps)} procedure, see~\cite{zhou2015cnnlocalization} for 
details. 
%
GAP also includes a thresholding function: given a score map for a class $y$, 
all locations that have a with a score larger than 20\% of 
the maximum score are selected, \ie 
%
        $D^y(X) = \{u | S^y_u(X) > 0.2 \max_{u \in \mathcal{U}} S^y_u(X)\}$.
%
At test time, for any input image five bounding boxes are produced, one 
for each class of the set of top-5 classes predicted by the convolutional 
neural network.
For each predicted class the bounding box is produced, so that it 
covers the largest connected component of $D^y(X)$\footnote{Better
results can be achieved by using a more involved method based on 
multiple image crops. This was used for the results 
in~\cite{zhou2015cnnlocalization}, but is not described in the 
manuscript, so we use the simpler but reproducible setting.}
%
%
Trained on the 1.2 million ILSVRC \emph{train} images, this approach 
achieves an error rate of 49.9\% on the ILSVRC \emph{val} set. 
%

\medskip\noindent\textbf{Improving bounding box predictions.}
We use the proposed micro-annotation technique to improve the \emph{CAM} localization score maps.
For computing mid-level feature representations (patterns) we use the \emph{conv5\_3} layer of the modified 
VGG~network from \cite{zhou2015cnnlocalization}.
This choice is motivated by the closely related papers~\cite{zhou2014object,simon2015neural} that studied object/part detectors 
emerging in convolutional neural networks.
We set $\rho=0.7$ as clustering threshold parameter and predict between $2$ and $4$ clusters. 
%
For the majority of classes (all except 56) we obtain only two clusters. 
%

\begin{figure}[t]
    \center
    \includegraphics[width=1.0\textwidth]{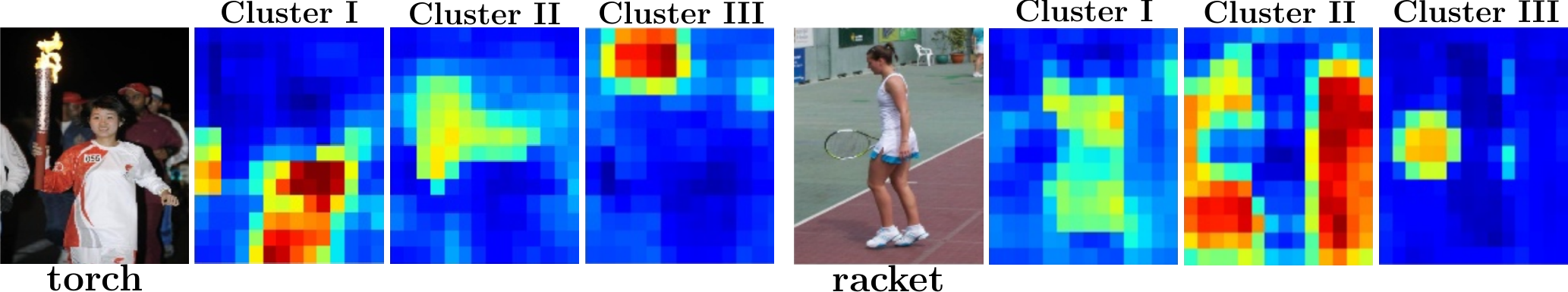}
    \caption{Examples of mid-level pattern clusters:
             \emph{torch} has clusters that correspond to \emph{person}, \emph{torch} and \emph{fire},
             and \emph{racket} has clusters that correspond to \emph{player}, \emph{court} and \emph{racket}.
             }\label{fig:many}
\end{figure}

Annotating all clusters for the 1000 classes requires less then 6 hours of annotator time.
In 182 semantic classes at least one cluster was identified as \emph{distractor}, while 
for the remaining classes different clusters typically correspond to different object 
parts. See Figure~\ref{fig:many} for a visualization of obtained object parts. 

\begin{figure}[t]
    \center
    \includegraphics[width=1.0\textwidth]{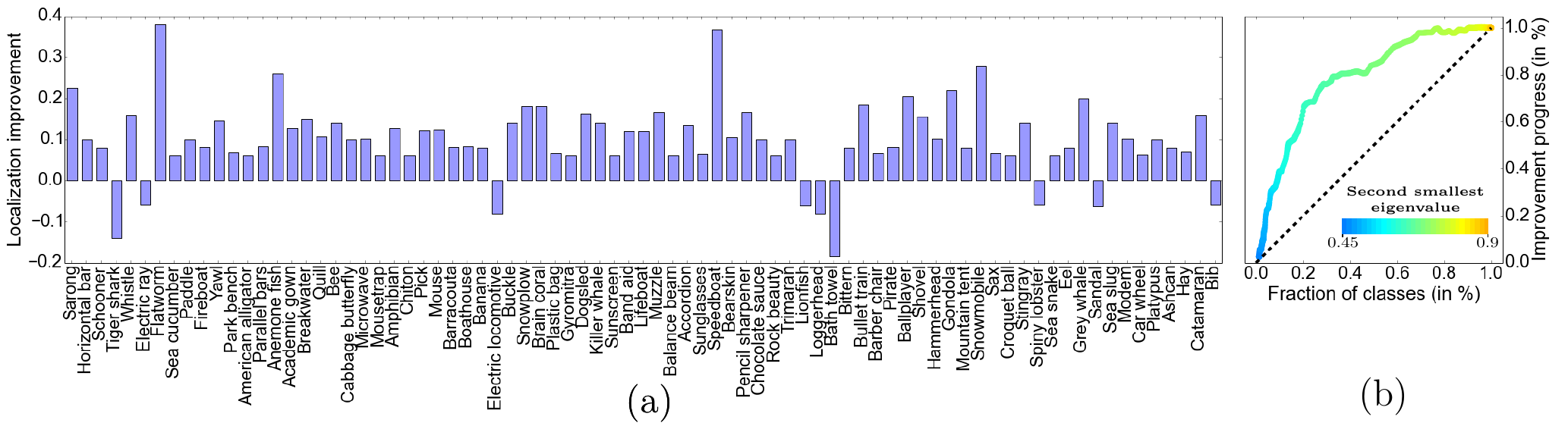}
    \caption{(a) Improvement of the localization scores by applying micro-annotation technique (for 79 classes with biggest changes);
             (b) Visualization of the trade off between the fraction of annotated classes and the fraction of overall improvement.}\label{fig:loc-results}
\end{figure}

After modifying the localization scores according to our method we compute bounding boxes
for all \emph{val} images. 
By construction, only the localization scores for 182 classes are affected compared 
to the baseline GAP results. 
On average, the localization performance improved by 4.9\% for those classes,
with individual improvements up to 38\% (\emph{flatworm}).
In Figure~\ref{fig:loc-results}(a) we illustrate the results 
for the 79 classes for which localization score changed by 
at least 5\% compared to the baseline.
We observed that for vast majority of classes our method helps to improve 
localization performance. 
As expected, many of these are examples where object and background 
consistently co-occur, for example \emph{speedboat} (improved from 42\% to 80\%) 
or \emph{snowmobile} (improved from 32\% to 60\%). 
For a few classes, we observed a decrease of performance. 
We inspected these visually and observed a few possible reasons: 
one possibility is that distractors are present in the image, but 
they actually help to find a better bounding box: for example, drawing 
a bounding box around a complete person often achieves above 50\% 
intersection-over-union for predicting \emph{bath towels}. 
A second possibility is that objects consist of multiple visually 
disconnected parts, \eg \emph{sandals}. GAP's large-connected-component 
rule tends to fail for these, but by including distractors into 
the score map, such as the \emph{foot}, can accidentally overcome 
this issue. 
%
%
We believe that this insight will be helpful for designing future 
weakly-supervised localization methods. 


%

We additionally investigate the question whether the eigenvalues, 
which are produced by spectral clustering, can be leveraged to
automatically identify classes with distractors.
For this purpose we sort all 1000 classes by the second smallest 
eigenvalue given by spectral clustering in increasing order
and study how the cumulative improvement of localization quality 
varies when micro-annotation is collected only for parts of the classes.
Figure~\ref{fig:loc-results}(b) depicts the curve, with the fraction 
of annotated classes on the $x$-axis and the fraction of 
localization improvement on the $y$-axis. 
One can see that the eigenvalues may be used to efficiently trade 
off annotation effort for localization performance.
For example, it is sufficient to annotate 15\% of all classes
to obtain 50\% of overall localization improvement,
or 50\% of the all classes for 85\% of improvement.

\begin{figure}
        \center
        \includegraphics[width=0.9\textwidth,height=0.28\textheight]{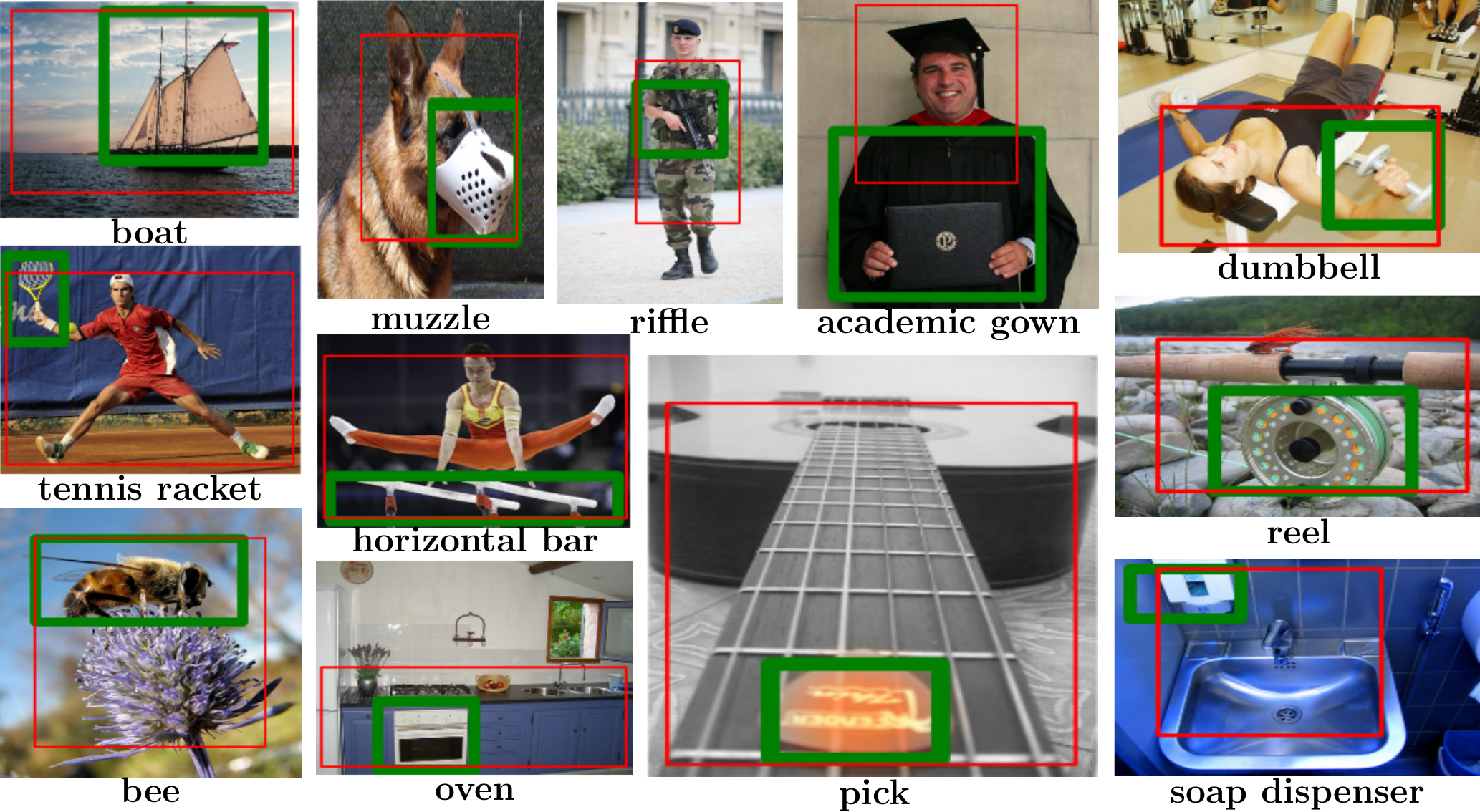}
        \caption{Typical mistakes (red boxes) of the baseline approach on ILSVRC \emph{val} that are corrected by our method (green boxes).
                 In particular, we demonstrate the following cases of foreground/background confusion:
                 \emph{boat/water}, \emph{muzzle/dog}, \emph{riffle/soldier},
                 \emph{academic gown/academic hat}, \emph{dumbbell/sportsman}, \emph{tennis racket/tennis player},
                 \emph{horizontal bar/gymnast}, \emph{pick/guitar}, \emph{reel/rod}, \emph{bee/flower}, \emph{oven/kitchen},
                 \emph{soap dispenser/shell}.}
\end{figure}

\begin{figure}[t]
        \center
        \includegraphics[width=0.45\textwidth]{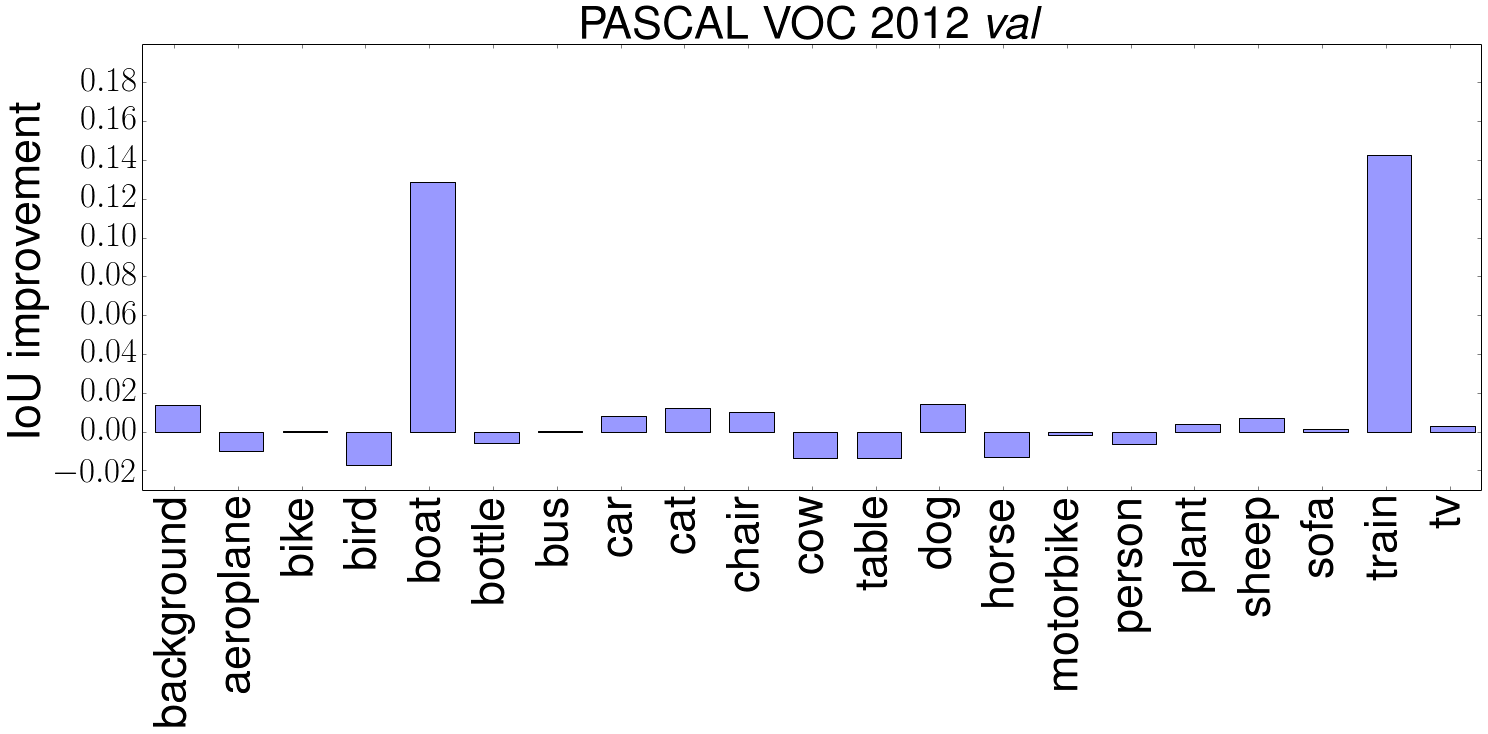}
        \quad
        \includegraphics[width=0.45\textwidth]{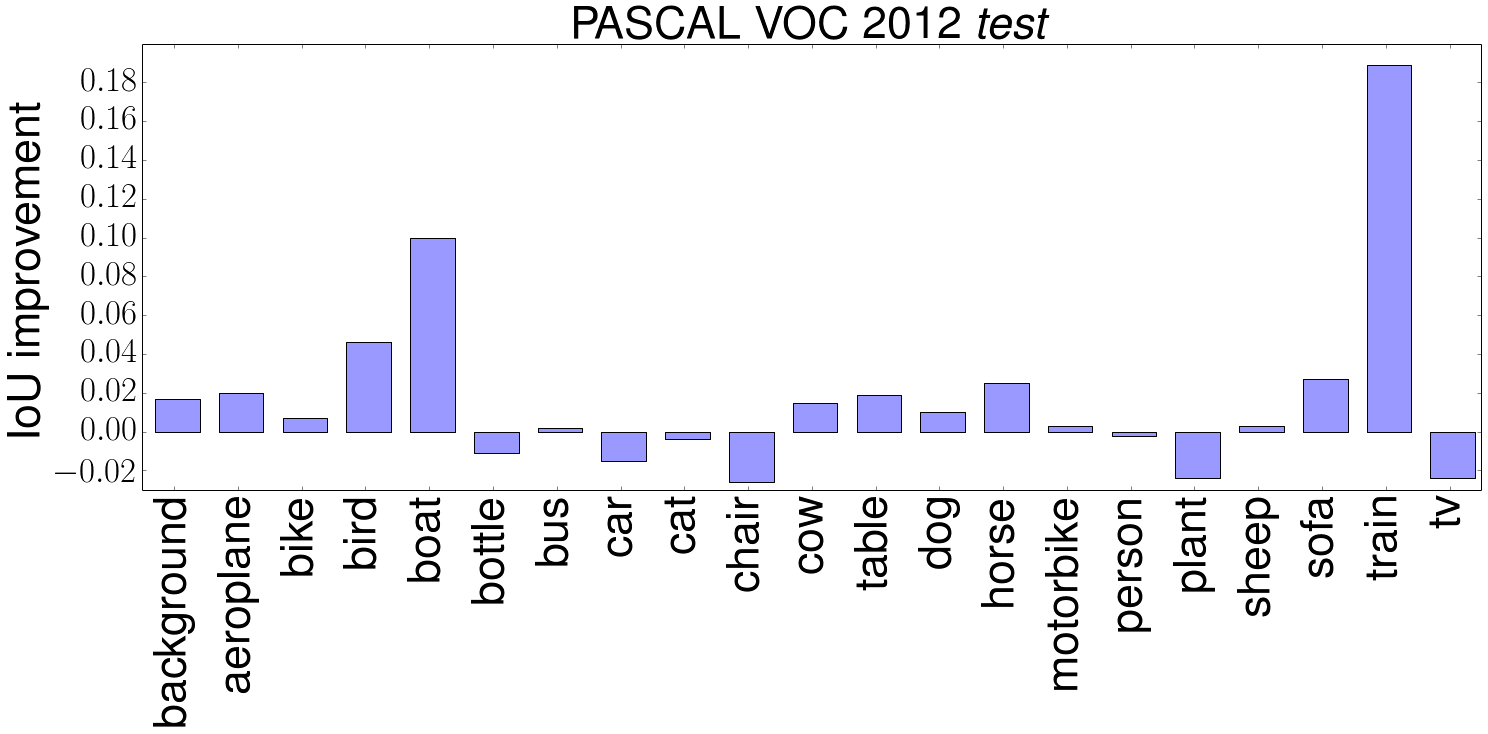}
        \caption{Improvement of the intersection-over-union scores by applying micro-annotation.}\label{fig:seg-results}
\end{figure}

\begin{figure}
        \center
        \includegraphics[width=0.85\textwidth]{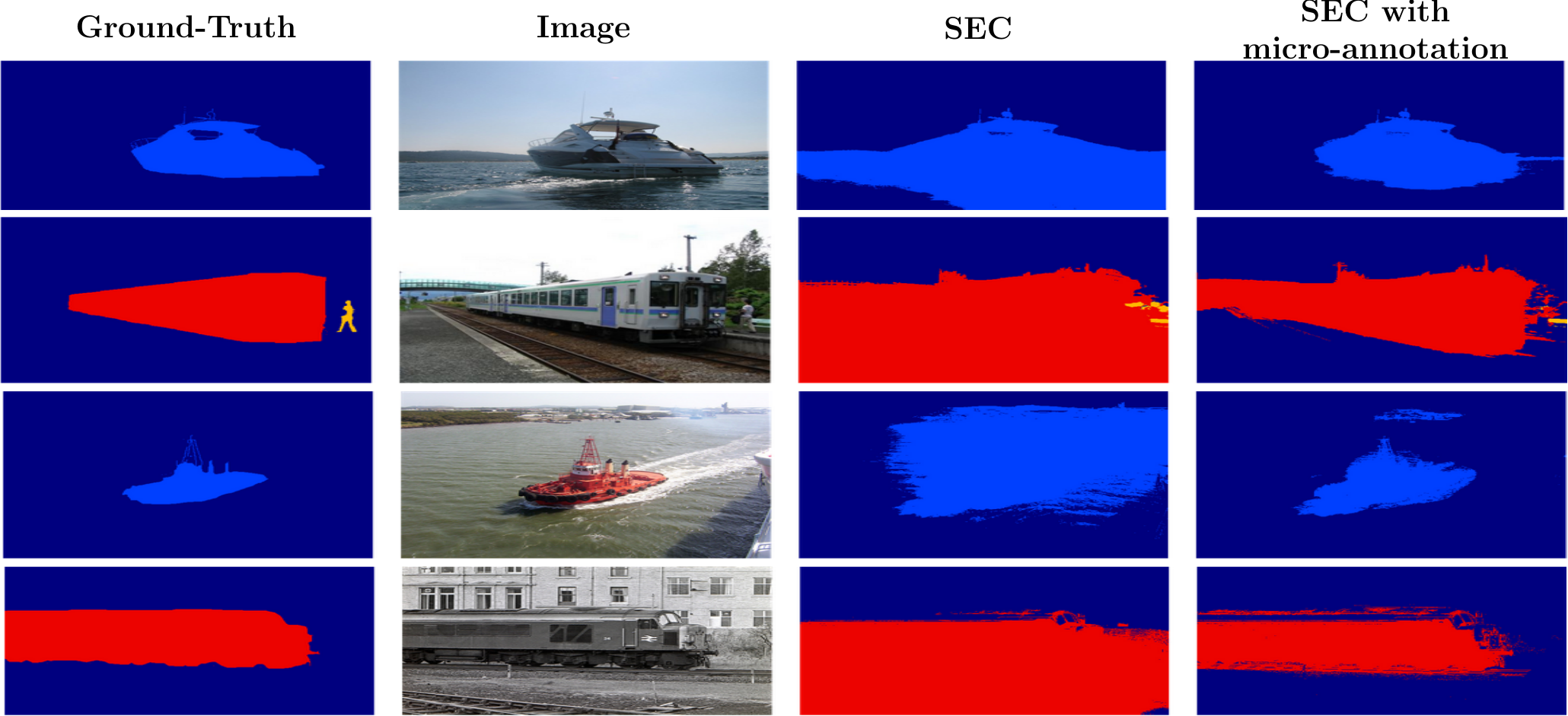}
        \caption{Examples of predicted segmentations masks without and with micro-annotation.}\label{fig:seg-examples}
\end{figure}

\medskip\noindent\textbf{Semantic image segmentation.}
Micro-annotation can also improve weakly-supervised semantic image segmentation.
We follow the protocol of the \emph{PASCAL VOC} challenges~\cite{everingham2010pascal}: the 
goal is to produce segmentation masks by assigning one of 21 labels 
(20 semantic classes or background) to each pixel of an image. 
The evaluation metric is the mean intersection-over-union scores 
across all labels.

We are again interested in the weakly-supervised setting where 
models are trained only from per-image class annotation.
Following the common practice, we use the \emph{PASCAL VOC2012} data
with augmentation from~\cite{BharathICCV2011}. In total, the 
training set (\emph{train}) has 10,582 images. We report 
results on the validation part (\emph{val}, 1449 images), 
and the test part (\emph{test}, 1456 images).
Because the ground truth annotations for the \emph{test} set 
are not public, we rely on the independent evaluation server\footnote{\url{http://host.robots.ox.ac.uk:8080/}} 
to obtain numerical results for this data.

The state-of-the-art technique for weakly-supervised image segmentation 
from image level labels is \emph{SEC}~\cite{kolesnikov2016seed}. Internally, it relies 
on \emph{CAM} localization score maps, as introduced in the previous section,
also based on a modified VGG-16 network~\cite{simonyan2014very} (but with 
different modifications), see~\cite{kolesnikov2016seed} for details.
In its original form, SEC achieves average intersection-over-union
scores of 50.7\% (val) and 51.5\% (test).

\medskip\noindent\textbf{Improving semantic image segmentation.}
We obtain mid-level pattern clusters for the 20 semantic classes 
of \emph{PASCAL VOC} following the protocol as in the previous
section. 
After annotating the clusters (which requires just a few minutes)
we found two classes that have significant distractors: \emph{boat} and \emph{train}.
Thus, we apply our method for these classes and retrain the \emph{SEC} model using the improved localization score maps. 
The per-class numerical evaluation for the \emph{val} and \emph{test} sets is presented in Figure~\ref{fig:seg-results}.
We observe significant improvement of the intersection-over-union metric
for the \emph{boat} class (12.9\% on \emph{val}, 14.2\% on \emph{test})
and
for the \emph{train} class (10.0\% on \emph{val}, 18.9\% on \emph{test}).
The performance for the other classes does not change significantly, only small 
perturbations occur due to the shared feature representation learned by the deep network. 
Overall, we achieve 51.9\% and 53.2\% mean intersection-over-union for the \emph{val} and \emph{test} sets,
and improvement of 1.3\% and 1.8\% percent over the original \emph{SEC} method.
For a visual comparison of the segmentations predicted by the baseline and our approach see Figure~\ref{fig:seg-examples}.

\medskip\noindent\textbf{Reproducibility.}
We implemented the proposed method in \texttt{python} using the \emph{caffe}~\cite{jia2014caffe} deep learning framework.
We will publish the pretrained models and code when the BMVC anonymity requirements are lifted.

\section{Conclusion}

Weakly-supervised localization techniques have the inherent problem 
of confusing objects of interest with consistently co-occuring 
distractors.
In this paper we present a micro-annotation technique that 
substantially alleviates this problem.
Our key insight is that objects and distractors can be distinguished
from each other because they form different clusters in the 
distributed representation learned by a deep 
network. 
We derive an annotation technique that requires only 
a few mouse clicks of user interaction per class and
we propose an algorithm for incorporating the acquired 
annotation into the localization procedure.

%
Experiments on the \emph{ILSVRC} 2014 and \emph{PASCAL} 2012 
demonstrate that the proposed micro-annotation method improves 
results further even for the state-of-the-art for weakly-supervised 
object localization and image segmentation.

\bibliography{bmvc}
\end{document}